\documentclass[10pt,twocolumn,letterpaper]{article}

\usepackage{cvpr}              % To produce the CAMERA-READY version
\usepackage{graphicx}
\usepackage{amsmath}
\usepackage{amssymb}
\usepackage{booktabs}
\usepackage{multirow}
\usepackage{multicol}
\usepackage{comment}
\usepackage{CJKutf8}
\usepackage[pagebackref,breaklinks,colorlinks]{hyperref}

\usepackage[capitalize]{cleveref}
\crefname{section}{Sec.}{Secs.}
\Crefname{section}{Section}{Sections}
\Crefname{table}{Table}{Tables}
\crefname{table}{Tab.}{Tabs.}

\def\confName{CVPR}
\def\confYear{2022}

\begin{document}

\title{Towards Learning Effective Visual-language Representations for \\Sign Language Translation}
\title{A Simple Multi-Modality Transfer Learning Baseline for \\ Sign Language Translation}

\author{Yutong Chen$^{1}$\thanks{Accomplished during Yutong Chen’s internship at MSRA.}  \quad Fangyun Wei$^{2}$\quad Xiao Sun$^{2}$ \quad Zhirong Wu$^{2}$ \quad Stephen Lin$^2$ \vspace{4pt}\\
	$^1$Tsinghua University \quad
    $^2$Microsoft Research Asia \\
    {\tt\small chenytjudy@gmail.com} \quad
	{\tt\small \{fawe, xias, wuzhiron, stevelin\}@microsoft.com} \\
}

\maketitle

\begin{abstract}
This paper proposes a simple transfer learning baseline for sign language translation.
Existing sign language datasets (e.g. PHOENIX-2014T, CSL-Daily) contain only about 10K-20K pairs of sign videos, gloss annotations and texts, which are an order of magnitude smaller than typical parallel data for training spoken language translation models.
Data is thus a bottleneck for training effective sign language translation models.
To mitigate this problem, we propose to progressively pretrain the model from general-domain datasets that include a large amount of external supervision to within-domain datasets. Concretely, we pretrain the sign-to-gloss visual network on the general domain of human actions and the within-domain of a sign-to-gloss dataset, and pretrain the gloss-to-text translation network on the general domain of a multilingual corpus and the within-domain of a gloss-to-text corpus. 
The joint model is fine-tuned with an additional module named the visual-language mapper that connects the two networks. This simple baseline surpasses the previous state-of-the-art results on two sign language translation benchmarks, demonstrating the effectiveness of transfer learning.
With its simplicity and strong performance, this approach can serve as a solid baseline for future research. Code and models are available at: \url{https://github.com/FangyunWei/SLRT}.
\end{abstract} 
\vspace{-3mm}
\section{Introduction}
\label{sec:intro}

\begin{figure}[t]
  \centering
  \includegraphics[width=0.95\linewidth]{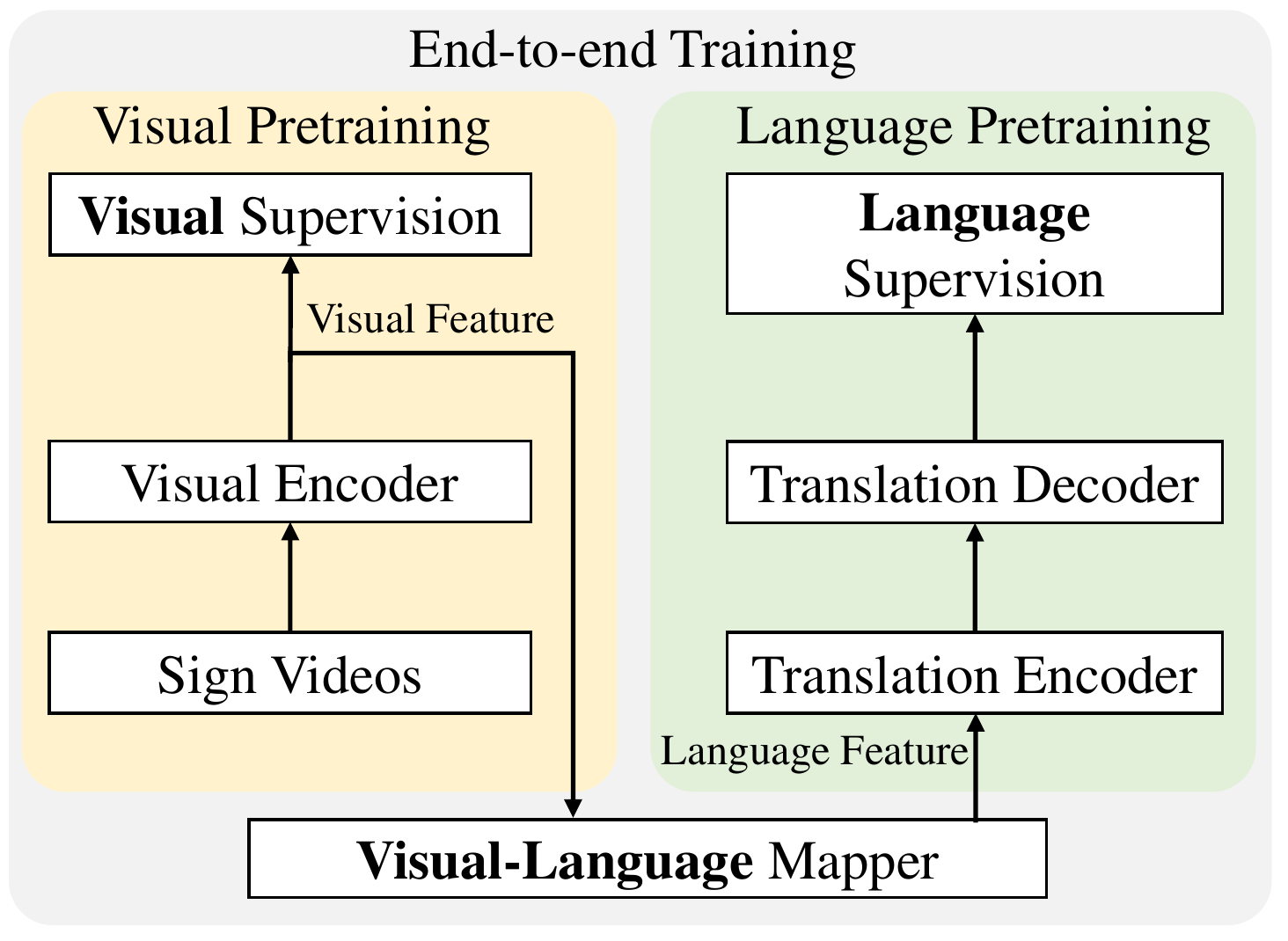}
  \vspace{-2mm}
   \caption{We decouple sign language translation into a visual task (left part) and a language task (right part), and propose a visual-language mapper (V-L Mapper) to bridge the connection between them. The decoupling allows both the visual and language networks to be effectively and independently pretrained before joint training. Both spatio-temporal information from sign videos and semantic knowledge from text transcriptions are encoded through VL-Mapper.}
   \label{fig:teaser}
   \vspace{-4mm}
\end{figure}

Sign languages are visual signals for communication among the deaf and hard of hearing. These languages are primarily expressed through manual articulations, but are also greatly aided by the movement of body, head, mouth, eyes and eyebrows. While technology for automatic machine translation of spoken languages have successfully been deployed in production \cite{vaswani2017attention,liu2020multilingual,XLM,xlm-r}, research on sign language translation (SLT) lags behind and is still in early-stage development. An effective system for automatic sign language translation may help to build a bridge between hearing-impaired and unimpaired people. 

Existing sign language translation methods follow the framework of neural machine translation (NMT) originally developed for spoken languages \cite{camgoz2018neural,Camg2020Multichannel,Yin2020STMCTransf,haozhou2020STMC,camgoz2020sign,zhou2021improving}, with the distinction that the source language is represented as spatio-temporal pixels instead of discrete tokens. 
To be concrete, sign videos are first fed into a video backbone network to extract an intermediate representation, which is then mapped to the target language text via NMT. The intermediate representation is usually supervised by glosses\footnote{Glosses are the word-for-word transcription of sign language where each gloss is a unique label for a sign. Typically, we identify each gloss by a capitalized word which is loosely associated with the sign's meaning.}~\cite{zhou2021improving,haozhou2020STMC,camgoz2020sign}, where each gloss corresponds to the semantic meaning of a single sign (e.g. happy, sad) in the continuous video input.

Despite adopting the formulation of advanced neural machine translation, the current results are far from satisfactory. The best reported sign language translation performance~\cite{zhou2021improving} on the PHOENIX-Weather-2014T test dataset~\cite{camgoz2018neural} is 24.32 in terms of BLEU-4, while a baseline transformer achieves a 30.9 BLEU-4 score for English to German translation \cite{liu2020multilingual}.
We hypothesize that the key factor that hinders the progress of sign language translation is the scale of the training data. To effectively train a typical NMT model, it usually requires a corpus of 1M paralleled samples~\cite{Sennrich2019Lowresource}. However, existing sign language datasets are an order of magnitude smaller, containing only fewer than 20K paralleled samples~\cite{camgoz2018neural,zhou2021improving}.

In this paper, we study a multi-modal pretraining approach to cope with the data scarcity issue for sign language translation. While pretraining and transfer learning has greatly improved performance in tasks of vision~\cite{kaiming2016cvpr,liu2021Swin,dosovitskiy2020vit}, language~\cite{BERT,Radford2019gpt,Mike2019BART,XLM,liu2020multilingual} and cross-modality~\cite{li2020oscar, zhang2021vinvl, Radford@Clip,Luo2020UniVL,lei2021less}, they are still under explored in SLT. Our work aims to exploit their strength in SLT.

SLT can be broken down into two disjoint tasks: a visual action recognition task that converts sign videos into semantic glosses (Sign2Gloss), and a language translation task that maps glosses into spoken language texts (Gloss2Text). Our transfer learning approach progressively pretrains each task separately and then finetunes the joint model.
For Sign2Gloss, we first pretrain the visual model on a general domain to learn generic human actions~\cite{K400_dataset,li2020word}, and then we transfer it to within the domain to learn fine-grained glosses.
Similarly for Gloss2Text, we adopt mBART~\cite{liu2020multilingual}, a denoising auto-encoder pretrained on a large-scale general-domain multilingual corpus, and transfer it to the within-domain task of gloss-to-text translation. By leveraging existing datasets and supervisions that can effectively transfer to sign language translation, the necessity of gathering large parallel data is lessened. 

With well-trained Sign2Gloss and Gloss2Text modules, we can build a two-staged pipeline known as Sign2Gloss2Text to generate a gloss sequence from the video and then translate the predicted gloss sequence into text. This two-staged pipeline is also implemented in~\cite{camgoz2018neural,camgoz2020sign,Yin2020STMCTransf,zhou2021improving} and shows promising results. However, glosses are discrete representations of the language modality, without encoding any spatio-temporal visual information from sign videos such as facial expressions\footnote{In sign languages, facial expressions are used to express both linguistic information and emotions.}, which may lead to degraded translation performance. For example, hearing-impaired individuals use exaggerated facial expressions to convey the 
adverb `Extremely', but this kind of information is ignored in gloss annotation. In contrast, labelers and linguists have to take into account these adverbs to produce translated sentences that are complete and semantically accurate. Incorporation of both visual and language modalities is thus needed.

To this end, we introduce a visual-language mapper which connects the visual features before gloss classification in the visual model to the gloss embedding in the translation model. With this mapper, the full model is jointly optimized and the discrete gloss representation is circumvented in joint training. The mapper is simply implemented as a fully connected MLP with two hidden layers. Figure~\ref{fig:teaser} shows our design.

In contrast to previous works which attempt to improve translation performance by integrating multiple cues from mouthing or pose in a handcrafted manner~\cite{Camg2020Multichannel,haozhou2020STMC} or by adopting advanced machine translation techniques such as back-translation~\cite{zhou2021improving}, our overall framework is extremely simple, resulting in a transfer learning approach on top of a standard NMT model. Some previous works conduct transfer learning for SLT by pretraining visual backbone on human action recognition~\cite{li2020tspnet} or loading pretrained word embeddings~\cite{li2020tspnet, Yin2020STMCTransf}, while we are the first to adopt both general-domain and within-domain pretraining in a progressive manner and incorporate pretrained spoken language model into SLT. Our experimental results demonstrate that this progressive pretraining of visual and translation models greatly boosts performance.
Our simple approach surpasses all existing methods by large margins, including those that employ semi-supervised learning, on PHOENIX-2014T \cite{camgoz2018neural} and CSL-Daily \cite{zhou2021improving}.
\section{Related Work}
\label{sec:related}
\noindent\textbf{Sign Language Recognition.}
A fundamental task in sign language understanding is Isolated Sign Language Recognition (ISLR), which aims to identify a single gloss word label for a short video clip \cite{MSASL,li2020word,imashev2020krsl,autsl,li2020transferring, Albanie2020bsl1k}. The more challenging task of Continuous Sign Language Recognition (CSLR) seeks to convert a continuous sign video into a gloss sequence using only weak sentence-level annotations \cite{Cui2017Recurrent,haozhou2020STMC,koller2019weak,koller2017re-sign,Pu2019Iterative}. Our work fully exploits gloss annotations for SLT by transferring within-domain knowledge from ISLR to CSLR and SLT.

\begin{figure*}[th!]
    \centering
    \includegraphics[width=0.98\textwidth]{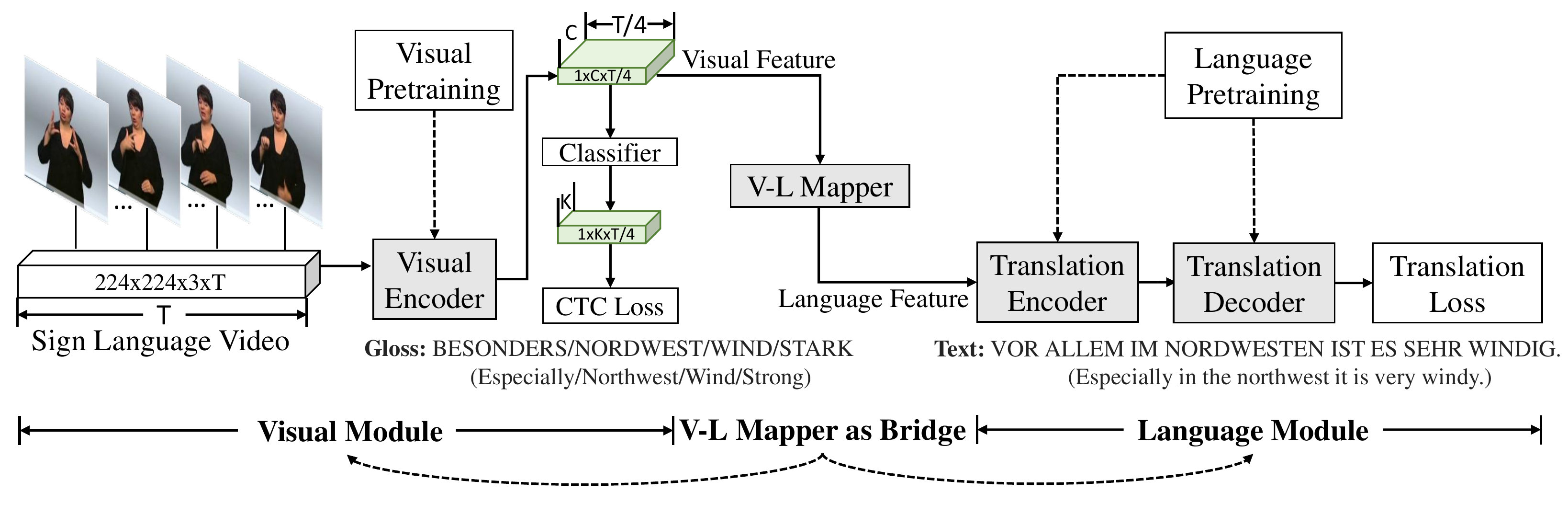}
    \vspace{-2mm}
    \caption{Overview of our framework. We decouple sign language translation into a visual task and a language task. The proposed visual-language mapper (V-L Mapper) establishes a bridge between features of the visual modality and language modality for end-to-end training. The decoupling allows both visual and language networks to be progressively and independently pretrained. }
    \label{fig:framework}
    \vspace{-5mm}
\end{figure*}

\noindent\textbf{Sign Language Translation.}
Sign Language Translation (SLT) aims to translate a raw video sequence to a spoken language sentence \cite{camgoz2018neural, camgoz2020sign, Camg2020Multichannel, haozhou2020STMC, zhou2021improving, Duarte2021how2,li2020tspnet,Yin2020STMCTransf}. Existing works attempt to formulate this task as a neural machine translation (NMT) problem. 
However, unlike NMT which benefits from a large-scale parallel corpus, SLT greatly suffers from data scarcity. To tackle this issue, \cite{camgoz2020sign} jointly trains SLR and SLT to enforce regularization on the translation encoder; \cite{zhou2021improving} proposes a data augmentation strategy of back-translating text to visual features using glosses as the pivot. Moreover, \cite{Camg2020Multichannel,haozhou2020STMC} manually design sophisticated multi-cue channels to model the collaboration of multiple visual cues in sign language, and \cite{li2020tspnet} introduces a temporal semantic pyramid network to capture multiple levels of temporal granularity in sign videos. Compared to these efforts, our method is simple yet more effective by utilizing a large amount of external supervision through progressive pretraining.

\noindent\textbf{Action Recognition.}
A related research field that may facilitate visual modeling of sign language is action recognition, where many works focus on network architecture \cite{qiu2017learning, xie2018rethinking, fan2020pyslowfast, Feichtenhofer2020X3D, carreira2017quo} and large-scale dataset construction \cite{K400_dataset, Raghav2017Sth, Fabian2015Act}. As fine-grained gesture understanding is a special case of human action recognition, some works for ISLR~\cite{Albanie2020bsl1k,li2020transferring,li2020word} and SLT~\cite{li2020tspnet} initialize their visual network with weights pretrained on action classification task. We employ general-domain pretraining on action recognition together with within-domain pretraining on Sign2Gloss in a progressive manner.

\noindent\textbf{Pretraining for Text Generation.}
Recently, the NLP community has seen rapid progress in large-scale self-supervised pretraining \cite{BERT,Mike2019BART,Raffel2020T5,Radford2019gpt,xlm-r}, which brings significant gains on downstream tasks. In particular, pretraining a language model on a large-scale monolingual corpus brings large improvements in low-resource NMT \cite{liu2020multilingual, xlm-r, MASS, baziotis2020languageprior}. Some multi-modality tasks such as image caption and VQA also leverage pretrained language models as initialization for bi-modal transformers~\cite{li2020oscar, Luo2020UniVL, VinVL_2021}. As sign language is a full-fledged language system, powerful NLP techniques can likely be extended into SLT to help address the data-scarcity issue. We are the first to apply a pretrained language model for spoken language in SLT.

\noindent\textbf{Transfer learning in SLT.}
Some previous works attempt to transfer external vision or language knowledge to SLT. For visual pretraining, \cite{li2020tspnet} pretrains the visual backbone on Kinetics-400~\cite{K400_dataset} and two ISLR datasets~\cite{MSASL,li2020word}. ~\cite{camgoz2018neural, camgoz2020sign, zhou2021improving,haozhou2020STMC} pretrain their visual backbones on within-domain Sign2Gloss task with gloss annotations. We adopt both general-domain and within-domain pretraining in a progressive manner. For language pretraining, ~\cite{li2020tspnet, Yin2020STMCTransf} load pretrained word embeddings into the decoder embedding layer but fail to demonstrate their effectiveness. We are the first to leverage powerful pretrained language models, which brings significant improvement.
\vspace{-2mm}
\section{Method}
\label{sec:method}
\vspace{-2mm}

In this section, we introduce our simple method for sign language translation. Given an input sign video $\mathcal{V}=(v_1,...,v_T)$ with $T$ frames, our goal is to learn a neural network $N_\theta(\cdot)$ that can predict the associated spoken language sentence $\mathcal{S}=(s_1,...,s_U)$ with $U$ words directly from the sign video $\mathcal{V}$:
\begin{equation}
    \mathcal{S} = N_\theta(\mathcal{V}).
    \vspace{-1mm}
\end{equation}
In order to transfer knowledge from general domains of action recognition and machine translation, we break down the SLT framework into two disjoint tasks: a visual action recognition task that converts the sign videos to semantic glosses (Sign2Gloss), and a language translation task that maps glosses to spoken language texts (Gloss2Text). This allows us to pretrain each task separately and then finetune the joint model.

In our approach, the overall network $N_\theta(.) $ is composed of three sub-networks: 1) a visual encoder network $\mathcal{E}$ that transforms raw video into visual features; 2) a sequence-to-sequence translation network $\mathcal{D}$ that translates language features into spoken language text; 3) a visual-language mapper $\mathcal{M}$ that bridges between features of the visual modality and language modality for joint training. The framework is illustrated in Figure~\ref{fig:framework}.

In this work, we demonstrate that using such a simple, no-frills framework can achieve high sign language translation performance. Besides its simplicity and high performance, we reveal that the bottleneck of current SLT systems mainly lies in the lack of training data, so that a more flexible architecture that can leverage as much training data as possible via pretraining, from both the vision and language sides, is superior.

\vspace{-2mm}
\subsection{Visual Encoder Network and Pretraining}
\label{sec:visual_encoder}
\vspace{-1mm}

\begin{figure}[t!]
    \centering
    \includegraphics[width=0.95\linewidth]{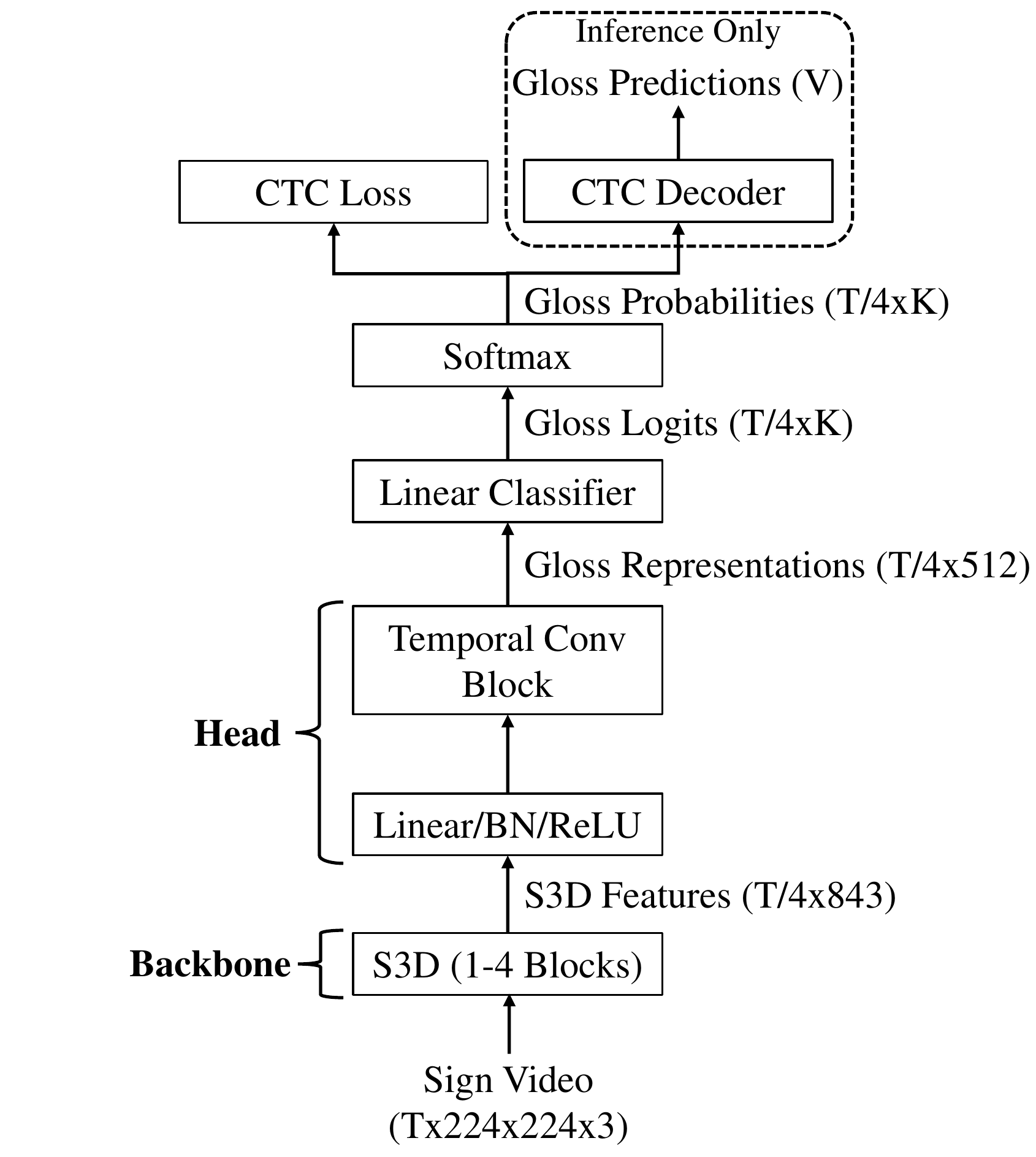}
    \vspace{-2mm}
    \caption{Architecture of our visual encoder network.}
    \label{fig:visual_encoder}
    \vspace{-5mm}
\end{figure}

The visual encoder network $\mathcal{E}$ transforms the raw video input into a visual feature. The visual feature in this stage is mainly used to predict gloss labels, which is essentially a fine-grained action recognition task. Figure~\ref{fig:visual_encoder} shows the network architecture, which consists of a video backbone and a lightweight head to further encode temporal information.

\noindent\textbf{Video Backbone.} We use S3D~\cite{xie2018rethinking} as our backbone due to its excellent trade-off between performance and inference speed. 
We feed each $T \times 224 \times 224 \times 3$ video into the backbone. Only the first four blocks of S3D are used since our goal is to extract a dense representation for gloss sequence prediction, and thus the extracted S3D features are of size $T/4 \times 832$ after spatial pooling. The extracted features then serve as the input of our head network.

\noindent\textbf{Head Network.} As shown in Figure~\ref{fig:visual_encoder}, our lightweight head network consists of a projection block containing a temporal linear layer, a batch normalization layer, and a ReLU layer, as well as a temporal convolutional block which contains two temporal convolutional layers with a temporal kernel size of 3 and a stride of 1, a linear translation layer, and a ReLU layer. We feed the S3D features into the projection block and the following temporal convolutional block to generate $\mathcal{Z} \in \mathcal{R}^{T/4 \times 512}$. We call it a gloss representation since it represents gloss categories in the high dimensional space. Then a linear classifier and a Softmax function are applied to extract frame-level gloss probabilities $\mathcal{P} \in \mathcal{R}^{T/4 \times K} $, where $K$ is the size of the gloss vocabulary. 

\noindent\textbf{Progressive Pretraining.} We progressively pretrain the visual encoder $\mathcal{E}$ by first pretraining it in a general domain to learn generic human actions and then transferring it to the within-domain task of learning fine-grained glosses. Specifically, for general-domain pretraining, we pretrain our S3D backbone on Kinetics-400, an action recognition dataset~\cite{K400_dataset} and then WLASL, an isolated sign recognition dataset~\cite{li2020word}. Next, for within-domain pretraining, we train our visual encoder under the Sign2Gloss task supervised by the continuous gloss annotations provided in SLT datasets.

 Unlike spoken language texts, the continuous gloss annotations are chronologically consistent with the sign signals.
We utilize the well-known connectionist temporal classification (CTC) loss~\cite{graves2006connectionist} for within-domain pretraining under the supervision of gloss annotations. The CTC loss considers all possible alignments between two sequences while minimizing the error. 
Concretely, for an input video $\mathcal{V}$ and the corresponding ground truth gloss sequence $\mathcal{G}$, we use CTC to compute $p(\mathcal{G}|\mathcal{V})$ by marginalizing over all possible $\mathcal{V}$ to $\mathcal{G}$ alignments:
\vspace{-1mm}
\begin{equation}
    p(\mathcal{G}|\mathcal{V}) = \sum_{\pi \in\mathcal{B}} p(\pi |\mathcal{V}),
    \vspace{-1mm}
\end{equation}
where $\pi$ denotes a path and $\mathcal{B}$ is the set of all viable paths that correspond to $\mathcal{G}$. The probability $p(\pi |\mathcal{V})$ is computed by the visual encoder $\mathcal{E}$. The CTC loss is then formulated as
\vspace{-1mm}
\begin{equation}
\label{eq:ctc}
    \mathcal{L} = - \ln p(\mathcal{G}|\mathcal{V}).
\end{equation}
\noindent\textbf{Gloss Sequence Prediction.} 
Once the pretraining is finished, our visual encoder network can be used to predict gloss sequences given sign videos. As shown in Figure~\ref{fig:visual_encoder}, we first use our visual encoder to extract gloss probabilities, then CTC decoding is utilized to generate the predicted gloss sequence. Details of the CTC decoding can be found in the supplementary materials. 
\vspace{-1mm}
\subsection{Translation Network and Pretraining}
\vspace{-1mm}
\label{sec:translationnetwork}
Now we introduce the translation network $\mathcal{D}$, which learns a mapping between gloss sequences and spoken language texts, and present the corresponding progressive pretraining procedure. 

\noindent\textbf{Translation Network.} Inspired by the recent progress of neural machine translation and multilingual denoising pre-training, we use mBART~\cite{liu2020multilingual}, a sequence-to-sequence denoising auto-encoder pretrained on large-scale multi-lingual corpus, as initialization for our translation network. The architecture is a standard sequence-to-sequence Transformer\cite{vaswani2017attention} with 12 layers for the encoder, 12 layers for the decoder, and a model dimension of 1024 on 16 heads.

\noindent\textbf{Progressive Pretraining.} 
With mBART initialization, our translation network is already pretrained in the general language domain. We further conduct within-domain pretraining on the Gloss2Text task to transfer mBART to the specific domain of gloss-to-text translation. Our goal is to train a translation network that can predict a text sentence $\mathcal{S}$ from a given gloss sequence $\mathcal{G}$. Concretely, we split both $\mathcal{G}$ and $\mathcal{S}$ into sub-word units using mBART's SentencePiece tokenizer \cite{SentencePiece} and project one-hot vectors into dense embeddings via mBART's pretrained word embedding layer. Then, we add positional embeddings to the word embeddings as inputs to the bottoms of the encoder and decoder stacks. We train mBART on the Gloss2Text corpus to minimize the sequence-to-sequence cross-entropy loss $\mathcal{L} = -\log P(\mathcal{S}|\mathcal{G})$. After obtaining a well-trained translation model, we can predict spoken language sentences given gloss sequences. Translating from \textit{ground-truth} sign gloss sequences to spoken language texts (Gloss2Text) is regarded as a virtual upper bound for performance on the SLT task ~\cite{camgoz2018neural,camgoz2020sign}. The two-stage translation task that first utilizes a Sign2Gloss model (our visual encoder) to generate a gloss sequence and then feeds the \textit{predicted} gloss sequence to a well-trained Gloss2Text pipeline is known as Sign2Gloss2Text. However, using glosses as the intermediate representations may be sub-optimal since glosses cannot fully encode spatio-temporal visual information. To overcome this limitation, we bridge vision and language modalities via our V-L mapper for joint training.

\vspace{-1mm}
\subsection{End-to-end Sign Language Translation}
\vspace{-1mm}
So far, we have described the architectures and the pretraining processes of our visual encoder and translation network. Now we introduce the Visual-Language Mapper (V-L Mapper), which builds a connection between the two networks modeling different modalities for the purpose of joint training. Our V-L Mapper is simply implemented as a fully-connected MLP with two hidden layers. As shown in Figure~\ref{fig:framework}, it converts visual features extracted by the visual encoder to language features, which are subsequently taken as the input of the translation encoder. We study the effects of feeding different visual features into the V-L Mapper in Section~\ref{sec:exp_end2end}, and use gloss representations (see Figure~\ref{fig:visual_encoder}) as our default setting. Thanks to the V-L Mapper, our framework can be trained in an end-to-end manner, under the joint supervision of the CTC loss and translation loss. Surprisingly, our framework even outperforms the acknowledged upper bound, i.e., translating from ground-truth sign gloss sequences to spoken language texts by using a well-trained Gloss2Text model, on the RWTH-PHOENIX-Weather-2014T test set. This is because our framework encodes both spatio-temporal information from sign videos and semantic knowledge from text transcriptions, providing more clues compared with the Gloss2Text model of only the language modality.

\vspace{-1mm}
\section{Experiments}
\vspace{-1mm}
\label{sec:experiment}

\begin{table*}[t]
\centering
\resizebox{\linewidth}{!}{%
\begin{tabular}{l|c c c c c  | c c c c c }
\toprule
 & \multicolumn{5}{c|}{Dev} & \multicolumn{5}{c}{Test} \\ 
Sign2Gloss2Text (two-stage) & R & B1 & B2 & B3  & B4 & R & B1 & B2 & B3 & B4 \\
\midrule
SL-Luong \cite{camgoz2018neural} & 44.14 & 42.88 & 30.30 & 23.02 & 18.40 & 43.80 & 43.29 & 30.39 & 22.82 & 18.13 \\
SL-Transf\cite{camgoz2020sign} & - & 47.73 & 34.82 & 27.11 & 22.11 & - & 48.47 & 35.35 & 27.57 & 22.45 \\
BN-TIN-Transf \cite{zhou2021improving} & 47.83 & 47.72 & 34.78 & 26.94 & 21.86 & 47.98 & 47.74 & 35.27 & 27.59 & 22.54 \\
BN-TIN-Transf + BT* \cite{zhou2021improving} & 49.53 & 49.33 & 36.43 & 28.66 & 23.51 & 49.35 & 48.55 & 36.13 & 28.47 & 23.51 \\
STMC-Transf \cite{Yin2020STMCTransf} &46.31 & 48.27 & 35.20 & 27.47 & 22.47 & 46.77 & 48.73 &  36.53 & 29.03& 24.00 \\
\midrule
Ours &  \textbf{50.23} & \textbf{50.36} & \textbf{37.50} & \textbf{29.69} & \textbf{24.63} & \textbf{49.59} & \textbf{49.94} & \textbf{37.28} & \textbf{29.67} & \textbf{24.60} \\

\midrule \midrule
Sign2Text (end-to-end) & R & B1 & B2 & B3  & B4 & R & B1 & B2 & B3 & B4 \\ \midrule
SL-Luong$^{\dagger}$  \cite{camgoz2018neural} & 31.80 & 31.87 & 19.11 & 13.16 & 9.94 & 31.80 & 32.24 & 19.03 & 12.83 & 9.58 \\
TSPNet-Joint$^{\dagger}$  \cite{li2020tspnet} & - & - & - & - & - & 34.96 & 36.10 & 23.12 & 16.88 & 13.41 \\
SL-Transf\cite{camgoz2020sign} & - & 47.26 & 34.40 & 27.05 & 22.38 & - & 46.61 & 33.73 & 26.19 & 21.32 \\
STMC-T\cite{haozhou2020STMC} &48.24 & 47.60 & 36.43 & 29.18 & 24.09 & 46.65 & 46.98 & 36.09 & 28.70 & 23.65 \\
BN-TIN-Transf + BT* \cite{zhou2021improving} & 50.29 & 51.11 & 37.90 & 29.80 & 24.45 & 49.54 & 50.80 & 37.75 & 29.72 & 24.32 \\
\midrule
Ours &  \textbf{53.10} & \textbf{53.95} & \textbf{41.12} & \textbf{33.14} & \textbf{27.61} & \textbf{52.65} & \textbf{53.97} & \textbf{41.75} & \textbf{33.84} & \textbf{28.39} \\
\bottomrule
\end{tabular}}
\vspace{-2mm}
\caption{Comparison with state-of-the-art methods on PHOENIX-2014T. $^{\dagger}$ denotes methods without using gloss annotations. * denotes methods with semi-supervised learning. `R' represents ROUGE, and `B1' denotes BLEU-1, with the same for `B2-B4'. Our framework outperforms all methods by large margins.}
\vspace{-5mm}
\label{tab:cmp_sota_phoenix}
\end{table*}

\subsection{Datasets and Evaluation Metrics}
\noindent\textbf{RWTH-PHOENIX-Weather 2014T.} PHOENIX-2014T \cite{camgoz2018neural} is the most widely used benchmark for SLT in recent years \cite{camgoz2018neural,camgoz2020sign,haozhou2020STMC,Yin2020STMCTransf,zhou2021improving}. The parallel corpus is collected from weather forecast news of the German public TV station PHOENIX over three years, including 8k triplets of RGB sign language videos of nine signers performing German Sign Language (DGS), sentence-level gloss annotations, and German translations transcribed from the news speaker. It contains 7096, 519 and 642 video segments in train, dev and test splits, respectively. The vocabulary size is 1066 for sign glosses and 2887 for German text. We compare our method with state-of-the-art methods on both the dev set and the test set. All ablation studies are conducted on this dataset.

\noindent\textbf{CSL-Daily.} CSL-Daily \cite{zhou2021improving} is a recently published Chinese sign language (CSL) translation dataset recorded in a studio. It contains 20k triplets of (video, gloss, text) performed by ten different signers. The content contains topics such as family life, medical care, and school life. CSL-Daily contains 18401, 1077 and 1176 segments in train, dev and test splits. The vocabulary size is 2000 for sign glosses and 2343 for Chinese text. We compare our approach with state-of-the-art methods on both the dev set and test set.

\noindent\textbf{Evaluation Tasks.} We examine performance on the following tasks:
\vspace{-2mm}
\begin{itemize}
    \item \textbf{Sign2Gloss}: Predict the gloss sequence given raw video input. This task is also known as CSLR (Continuous Sign Language Recognition). This task is mainly used to evaluate our visual encoder.
    \vspace{-2mm}
    \item \textbf{Gloss2Text}:
    Translate a \textit{ground-truth} gloss sequence to text. Its results are generally regarded as an upper bound for the sign language translation task. We also use this task to evaluate our translation model.
    \vspace{-2mm}
    \item \textbf{Sign2Gloss2Text}: A two-stage pipeline where we first adopt a Sign2Gloss module to predict a gloss sequence and then translate the \textit{predicted} glosses to text by a Gloss2Text module. We use this to evaluate pipelines in which the visual encoder and the translation model are connected by the predicted gloss sequence.
    \vspace{-2mm}
    \item \textbf{Sign2Text}: Directly translate sign language video into text, which is our goal.
\end{itemize}
\vspace{-2mm}
Following previous work \cite{camgoz2018neural,camgoz2020sign,zhou2021improving,haozhou2020STMC}, we use Word Error Rate (WER) to evaluate Sign2Gloss, and ROUGE~\cite{rouge} and BLEU~\cite{bleu} to evaluate the other three tasks.

\vspace{-1mm}
\subsection{Implementation details}
\vspace{-1mm}

Our model is implemented in PyTorch. Details about all hyperparameters are given in the supplementary materials.

\noindent\textbf{Visual Encoder Pretraining.}
We progressively pretrain the visual encoder from general domain to within domain. First we pretrain the S3D backbone on two action recognition datasets sequentially, namely Kinetics-400~\cite{K400_dataset}, the most popular human action recognition dataset containing 400 action classes, and WLASL~\cite{li2020word}, a large-scale Word-Level American Sign Language video datasets containing 2000 isolated sign classes. The training procedure follows~\cite{xie2018rethinking}. Video clips are fed through five blocks in S3D backbone followed by a 3D average pooling layer and a linear classification layer to predict the action class. Next we conduct within-domain pretraining on Sign2Gloss task using the CTC loss (Eq.~\ref{eq:ctc}), where we only use the first four blocks of the pretrained S3D and spatially pool the S3D features to the size of $T/4\times832$ as inputs of our head network.

\noindent\textbf{Translation Pretraining.} 
For general-domain pretraining, we initialize our language model with the official release of mBART-large-cc25\footnote{https://huggingface.co/facebook/mbart-large-cc25} which is pretrained on CC25, a multi-lingual corpus of size 1300GB from Common Crawl\footnote{https://commoncrawl.org/} that covers 25 languages. We also try GPT2~\cite{Radford2019gpt} pretrained on 16GB German monolingual corpus. Unless otherwise specific, we use mBART by default.

\noindent\textbf{Joint Training.} We load the two independently pretrained modules as the initialization for joint training. The features before the linear classifier, i.e. gloss representation, are projected into vectors of 1024 dimension by the V-L Mapper and position embeddings are added to them to form inputs to the translation encoder. The whole network is trained under the joint supervision of the CTC loss and cross-entropy loss with both weights set to 1.0.

\vspace{-1.5mm}
\subsection{Comparison with State-of-the-art Methods}
\vspace{-2mm}
We compare our approach to state-of-the-art methods on PHOENIX-2014T and CSL-Daily, as shown in Table~\ref{tab:cmp_sota_phoenix} and Table~\ref{tab:cmp_sota_csl}. Without integrating multi-cue features~\cite{haozhou2020STMC,Camg2020Multichannel} nor advanced data augmentation strategies such as back translation~\cite{zhou2021improving}, our simple method significantly surpasses all counterparts on PHOENIX-2014T and CSL-Daily.

\begin{table*}[t]
\centering
\resizebox{\linewidth}{!}{%
\begin{tabular}{l | c c c c c | c c c c c }
\toprule
 & \multicolumn{5}{c|}{Dev} & \multicolumn{5}{c}{Test} \\ 
Sign2Gloss2Text (two-stage) & R & B1 & B2 & B3  & B4 & R & B1 & B2 & B3 & B4 \\
\midrule
SL-Luong \cite{camgoz2018neural} & 40.18 & 41.46 & 25.71 & 16.57 & 11.06 & 40.05 & 41.55 & 25.73 & 16.54 & 11.03 \\
SL-Transf\cite{camgoz2020sign} & 44.18 & 46.82 & 32.22 & 22.49 & 15.94 & 44.81 & 47.09 & 32.49 & 22.61 & 16.24 \\
BN-TIN-Transf \cite{zhou2021improving} & 44.21 & 46.61 & 32.11 & 22.44 & 15.93 & 44.78 & 46.85 & 32.37 & 22.57 & 16.25 \\
BN-TIN-Transf + BT* \cite{zhou2021improving} & 48.38 & \textbf{50.97} & 36.16 & 26.26 & 19.53 & 48.21 & \textbf{50.68} & 36.00 & 26.20 & 19.67 \\
\midrule
Ours &  \textbf{51.35} & 50.89 & \textbf{37.96} & \textbf{28.53} & \textbf{21.88} & \textbf{51.43} & 50.33 & \textbf{37.44} & \textbf{28.08} & \textbf{21.46} \\
\midrule
\midrule
Sign2Text (end-to-end) & R & B1 & B2 & B3  & B4 & R & B1 & B2 & B3 & B4 \\
SL-Luong$^{\dagger}$ \cite{camgoz2018neural} & 34.28 & 34.22 & 19.72 & 12.24 & 7.96 & 34.54 & 34.16 & 19.57 & 11.84 & 7.56 \\
SL-Transf \cite{camgoz2020sign} & 37.06 & 37.47 & 24.67 & 16.86 & 11.88 & 36.74 & 37.38 & 24.36 & 16.55 & 11.79 \\
BN-TIN-Transf \cite{zhou2021improving} & 37.29 & 40.66 & 26.56 & 18.06 & 12.73 & 37.67 & 40.74 & 26.96 & 18.48 & 13.19 \\
BN-TIN-Transf + BT* \cite{zhou2021improving} & 49.49 & 51.46 & 37.23 & 27.51 & 20.80 & 49.31 & 51.42 & 37.26 & 27.76 & 21.34 \\
\midrule
Ours & \textbf{53.38} & \textbf{53.81} & \textbf{40.84} & \textbf{31.29} & \textbf{24.42} & \textbf{53.25} & \textbf{53.31} & \textbf{40.41} & \textbf{30.87} & \textbf{23.92}  \\ 

\bottomrule
\end{tabular}}
\vspace{-2mm}
\caption{Comparison with state-of-the-art methods on CSL-Daily. $\dagger$ denotes methods without using gloss annotations. * denotes methods with semi-supervised learning. Without any tricks nor using manually generated extra data, 
our method surpasses the recent semi-supervised method BN-TIN-Transf + BT~\cite{zhou2021improving}.
}
\label{tab:cmp_sota_csl}
\vspace{-2mm}
\end{table*}

\begin{table*}[t]
    \centering
    \begin{tabular}{c c c  |c c  | c c c c c |c c c c c }
    \toprule
            \multicolumn{3}{c|}{\multirow{2}{*}{Pretraining}} & \multicolumn{2}{c|}{Sign2Gloss} & \multicolumn{10}{c}{Sign2Text}  \\   \cmidrule{4-15}
            & & & \multicolumn{1}{c}{Dev} & \multicolumn{1}{c|}{Test} & \multicolumn{5}{c|}{Dev} & \multicolumn{5}{c}{Test} \\
         K & WL & S2G & WER & WER & R & B1 &B2 & B3 &B4 & R&B1&B2&B3&B4\\ \midrule
         & & \checkmark & 27.25 & 28.06 & 51.97 & 52.99 & 40.42 & 32.59 &27.27 & 51.82 & 52.53 & 40.18 & 32.37 & 27.01\\ 
         \checkmark & & \checkmark  & 23.05 & 23.50 & \textbf{53.24} & \textbf{53.99} & \textbf{41.47} & \textbf{33.63} & \textbf{28.19} & 52.42 & 53.66 & 41.27 & 33.36 & 27.91 \\ 
         \checkmark & \checkmark  & \checkmark & \textbf{21.90} & \textbf{22.45} & 53.10 & 53.95 & 41.12 & 33.14 & 27.61 & \textbf{52.64} & \textbf{53.97} & \textbf{41.75} & \textbf{33.84} & \textbf{28.39} \\
         \checkmark & \checkmark  &   & - & - & 45.84 & 47.31 & 33.64 & 25.83 & 20.76 & 45.93 & 47.40 & 34.30 & 26.47 & 21.44\\         
     \bottomrule
    \end{tabular}
    \vspace{-2.5mm}
    \caption{Ablation study of visual encoder with different pretraining settings on the \textbf{PHOENIX Sign2Gloss} and \textbf{PHOENIX Sign2Text} tasks. K, WL, S2G denote pretraining on Kinetics-400, WLASL and Sign2Gloss, respectively.}
    \label{tab:S3D_pretrain}
    \vspace{-5.5mm}
\end{table*}

\vspace{-1.5mm}
\subsection{Ablation Study}
\vspace{-1.5mm}
\subsubsection{Pretraining of Visual Encoder}
\label{sec:visual_pretraining}
\vspace{-2.5mm}
Our visual encoder is pretrained in a progressive manner. We first study the effects of using different general-domain pretraining strategies:
\vspace{-2.5mm}
\begin{itemize}
\item \textbf{Scratch.} No general-domain pretraining is conducted. The S3D backbone is trained from scratch.
\vspace{-3mm}
\item \textbf{K-400.} General-domain pretraining is done on Kinetics-400~\cite{K400_dataset}, a large-scale action recognition set.
\vspace{-7mm}
\item \textbf{K-400$\longrightarrow$WLASL.} We further pretrain the K-400 pretrained S3D backbone on WLASL~\cite{li2020word}, a large-scale word-level sign language recognition dataset.
\end{itemize}
\vspace{-3mm}
We conduct within-domain Sign2Gloss pretraining on these pretrained models and report the effects on both the Sign2Gloss and Sign2Text tasks in Table~\ref{tab:S3D_pretrain}. The performance of Sign2Gloss directly reflects the effects of different general-domain pretrained models. Although K-400 is an action classification dataset, using the model pretrained on it as initialization nevertheless improves Sign2Gloss performance compared to the model trained from scratch, reducing the WER from 28.06 to 23.50 on the test set. Using K-400$\longrightarrow$WLASL as the initialization further boosts performance, achieving 22.45 WER on the test set. Though there exist differences between WLASL and PHOENIX-2014T, e.g., the former is proposed to solve isolated sign language recognition for American sign language while the latter aims to solve continuous sign language recognition for German sign language, general-domain pretraining on WLASL still learns relevant representations, e.g., low-level gesture features. As for Sign2Text, the gains of visual pretraining become narrowed, which suggests that learning favorable visual features is not the only determining factor for Sign2Text. For example, the translation model provides complementary information. In addition, to verify the importance of within-domain Sign2Gloss pretraining for Sign2Text, we load the visual encoder only pretrained on K-400 and WLASL into Sign2Text joint training. As the last column in Table~\ref{tab:S3D_pretrain} shows, skipping within-domain pretraining considerably hurts performance, reducing BLEU-4 by nearly 7 on both sets. We conclude that both general-domain and within-domain pretraining contribute to our method's high performance.

\begin{table*}[t]
    \centering
    
        \begin{tabular}{c| c c c c c |c c c c c }
    \toprule
          General-domain & \multicolumn{5}{c|}{Dev} & \multicolumn{5}{c}{Test}\\
          Pretraining &  R& B1 &B2 & B3 & B4 & R& B1 &B2 & B3 & B4 \\ \midrule
        GPT2 Scratch    &  44.45 & 45.93 & 31.96 & 24.17 & 19.40 & 43.57 & 44.62 & 31.27 & 23.65 & 18.96 \\
        mBart Scratch & 46.56 & 48.42 & 35.22 & 27.55 & 22.71 & 47.03 & 48.16 & 35.32 &  27.63 & 22.57 \\
        GPT2 w/ German Corpus  & 49.99 & 50.86 & 38.07 & 30.32 & 25.11 & 48.61 & 48.98 & 36.55 & 28.88 & 23.82 \\
        mBART w/ CC25 & \textbf{53.79} & \textbf{54.01} & \textbf{41.41} & \textbf{33.50} & \textbf{28.19} & \textbf{52.54} & \textbf{52.65} & \textbf{39.99} & \textbf{32.07} & \textbf{26.70} \\ 
     \bottomrule
    \end{tabular}
    \vspace{-2.5mm}
    \caption{Ablation study of general-domain language pretraining on the \textbf{PHOENIX Gloss2Text} task. We conduct within-domain pretraining on the Gloss2Text task to transfer from the general domain to the specific domain of gloss-to-text translation.}
    \label{tab:plm}
    \vspace{-2mm}
\end{table*}

\begin{table*}[t]
    \centering
    \begin{tabular}{c c|c c c c c | c c c c c }
    \toprule
       \multirow{2}{*}{CC25} & \multirow{2}{*}{Gloss2Text} & \multicolumn{5}{c|}{Dev} & \multicolumn{5}{c}{Test} \\  
        &   & R& B1 &B2 & B3 & B4 & R& B1 &B2 & B3 & B4 \\
        \midrule
        &   & 46.47 & 46.99 & 33.75 & 25.83 & 20.70 & 46.67 & 47.45 & 34.39 & 26.47 & 21.36 \\
         & \checkmark   & 48.77 & 49.46 & 35.98 & 27.98 & 22.89 & 47.65 & 49.01 & 36.16 & 28.33 & 23.28 \\
       \checkmark &   & 52.30 & 52.39 & 39.97 & 32.28 & 26.99 & 51.83 & 53.05 & 40.34 & 32.28 & 26.95 \\
        \checkmark & \checkmark  & \textbf{53.10} & \textbf{53.95} & \textbf{41.12} & \textbf{33.14} & \textbf{27.61} & \textbf{52.65} & \textbf{53.97} & \textbf{41.75} & \textbf{33.84} & \textbf{28.39} \\
    \bottomrule
    \end{tabular}
    \vspace{-2.5mm}
    \caption{Ablation study of mBART with different pretraining settings on the \textbf{PHOENIX Sign2Text} task. `CC25' denotes transformer with mBART initialization, `Gloss2Text' represents transformer pretrained on the Gloss2Text task.}
    \label{tab:lm_end2end}
    \vspace{-2mm}
\end{table*}

\begin{table*}[h!]
    \centering
    \begin{tabular}{l|  c c c c c | c c c c c }
    \toprule
\multirow{2}{*}{Method} & \multicolumn{5}{c|}{Dev} & \multicolumn{5}{c}{Test} \\ 
& R & B1 & B2 & B3  & B4 & R & B1 & B2 & B3 & B4 \\
        \midrule
      Gloss2Text &  53.79 & 54.01 & 41.41 & 33.50 & 28.19 & 52.54 & 52.65 & 39.99 & 32.07 & 26.70 \\ 
      Sign2Gloss2Text & 50.23 & 50.36 & 37.50 & 29.69 & 24.63 & 49.59 & 49.94 & 37.28 & 29.67  & 24.60 \\ 
      \midrule
      Sign2Text w/ Gloss Logits & 52.45 & 53.23 & 40.55 & 32.66  & 27.23 & \textbf{52.71} & 53.70 & 41.20 & 33.22 & 27.78 \\
      Sign2Text w/ Gloss Reps & \textbf{53.10} & \textbf{53.95} & \textbf{41.12} & \textbf{33.14} & \textbf{27.61} & 52.65 & \textbf{53.97} & \textbf{41.75} & \textbf{33.84} & \textbf{28.39} \\ 
     Sign2Text w/ S3D Features &  43.53 & 44.11 & 31.46 & 24.60  & 20.24 & 43.77 & 44.68 & 32.02 & 25.02 & 20.62 \\ 
     \bottomrule
    \end{tabular}
    \vspace{-2.5mm}
    \caption{Ablations on different visual features as the V-L Mapper input on \textbf{PHOENIX Sign2Text}.}
    \label{tab:bridge}
    \vspace{-5mm}
\end{table*}

\vspace{-5mm}
\subsubsection{Pretraining of Translation Model}
\vspace{-2mm}
\label{sec:language_pretraining}
SLT greatly suffers from the data scarcity issue. Recently, language pretraining has shown promising results in low-resource NMT~\cite{MASS,XLM,liu2020multilingual,xlm-r}, which inspires us to introduce language pretraining to SLT. 

\noindent\textbf{General-domain Pretraining Improves Gloss2Text.}
    We first experiment with two popular pretrained language models, namely mBART~\cite{liu2020multilingual} and GPT2~\cite{Radford2019gpt}, to verify the effects of using different architectures and different large-scale general-domain corpus, through direct evaluation on the PHOENIX Gloss2Text task. Table~\ref{tab:plm} shows the results. As baselines, we train two translation networks with the same architecture as mBART or GPT2 but with random initializations. mBART outperforms GPT2, suggesting that the encoder-decoder architecture and bidirectional attention of mBART makes it more suitable for Gloss2Text than GPT2 which only has a decoder with unidirectional attention. However, general-domain pretraining on large corpus improves both mBART and GPT2 on Gloss2Text and mBART pretrained on CC25 achieves the best performance. We use mBART for further experiments. Additionally, mBART is pretrained on the multilingual corpus and thus can be used as a generic pretraining model for various sign languages.

\vspace{-2mm}
\noindent\textbf{Progressive Pretraining Improves Sign2Text.}
We examine the effects of progressive pretraining of the translation model on the Sign2Text task, which is our final goal. Four pretraining settings are studied: 1) without pretraining; 2) pretraining on the Gloss2Text task; 3) pretraining on the CC25 corpus; 4) progressive pretraining, i.e., the translation model is first pretrained on CC25, then a further within-domain pretraining is conducted on the Gloss2Text task. For all settings, we use the same joint training process for Sign2Text. The results are shown in Table~\ref{tab:lm_end2end}. We use transformer of the same architecture without pretraining as our baseline. From the table we can observe that the translation model pretrained on the Gloss2Text task yields a slight improvement (+1.92 BLEU-4 on the test set). When pretraining on CC25, our method achieves 26.95 BLEU-4 on the test set, which demonstrates the importance of language pretraining on the large-scale corpus. The best result is achieved by progressive pretraining, which can be attributed to both general-domain pretraining on the large-scale corpus and domain alignment through within-domain pretraining (Gloss2Text) and the downstream task (Sign2Text).

\vspace{-5mm}
\subsubsection{Joint Multi-modality Training}
\label{sec:exp_end2end}
\vspace{-2mm}
At last, we study the effectiveness of our joint multi-modality Sign2Text training by bridging the two modalities via the V-L Mapper. The most straightforward approach is to build a two-stage translation pipeline, i.e. Sign2Gloss2Text, where predicted glosses serve as the intermediate state. As the discrete glosses cannot fully capture the semantics in sign video, here we study using different visual features as the inputs to the V-L Mapper. In an ablation study, we examine the three features shown in Figure~\ref{fig:visual_encoder}, namely gloss logits, gloss representations, and S3D features. Table \ref{tab:bridge} shows the results. Translating from a ground-truth gloss sequence to text (Gloss2Text) is generally regarded as an upper bound in SLT. Surprisingly, our joint Sign2Text training with gloss representations and with gloss logits outperforms not only Sign2Gloss2Text, but also the Gloss2Text upper bound, which demonstrates the effectiveness of our progressive pretraining and the proposed multi-modality transfer learning. 

\vspace{-2.5mm}
\section{Conclusion}
\label{sec:conlusion}
\vspace{-2mm}
We present a simple yet effective multi-modality transfer learning baseline for sign language translation. To alleviate the data scarcity issue, we exploit large-scale external knowledge from human action and spoken language by progressively pretraining visual and language modules from general domains to within the target domains. The two individually pretrained modules are then bridged via the Visual-Language Mapper for joint SLT training.
Experiments on two SLT datasets show that our approach outperforms all state-of-the-art methods. Our method can be applied to various sign languages.
In future work, we would like to use this framework to transfer more external knowledge into SLT for further improvement. We hope our simple baseline can facilitate future research in SLT and motivate more researchers to engage in this field.
\appendix

\section{Implementation Details}
We elaborate on our training and inference procedure in this section. Unless otherwise specified we use a batch size of 8, the Adam optimizer with a weight decay of 1e-3 and the cosine annealing scheduler.

\noindent\textbf{Visual Encoder Pretraining.}
For general-domain pretraining on K-400~\cite{K400_dataset} and WLASL~\cite{li2020word}, our training procedure follows~\cite{xie2018rethinking}. All of the five blocks in S3D backbone followed by a spatial average pooling layer and a linear classification layer are used in the general-domain pretraining. For within-domain pretraining on Sign2Gloss, we only use the first four blocks of the pretrained S3D and spatially pool the S3D features into size of $T/4\times832$ as inputs into our head network. The data augmentations for within-domain pretraining include temporally-consistent spatial random crop with range of [0.7-1.0] and frame-rate augmentations with range of [$\times0.5$-$\times 1.5$]. All frames are spatially resized to 224$\times$224 as inputs. We train our visual encoder for 80 epochs with an initial learning rate of 1e-3. During inference, we use all frames and resize them to 224$\times$224 without cropping. For gloss sequence prediction, we adopt CTC beam search decoder with widths ranging from one to ten and choose the width of the best performance on dev set for test set evaluation. 

\noindent\textbf{CTC Decoder.}
Once the within-domain pretraining (Sign2Gloss) of our visual encoder is finished, we can use it for gloss prediction. Concretely, given a sign video $\mathcal{V}=(v_1,...,v_T)$ with $T$ frames, the visual encoder $\mathcal{E}$ predicts gloss probabilities $\mathcal{P}=(p_1,...,p_{T/4})$ where $p_t$ represents the distribution of gloss probability at step $t$. CTC decoder uses $\mathcal{P}$ as the input to estimate the most confident gloss sequence using beam search decoding algorithm. More details can be found in~\cite{graves2006connectionist}.

\noindent\textbf{Translation Pretraining.}
For general-domain pretraining, we use the release of mBART-large-cc25\footnote{https://huggingface.co/facebook/mbart-large-cc25} as initialization. For within-domain pretraining, i.e., Gloss2Text task, we train the translation network with the cross-entropy loss for 80 epochs with an initial learning rate of 1e-5. We also use dropout of 0.3 and label smoothing of 0.2 to prevent overfitting. For memory efficiency, we prune the mBART word embedding by preserving words in the target language, i.e., Chinese for CSL-Daily~\cite{zhou2021improving} and German for PHOENIX2014T~\cite{camgoz2018neural}. The pruned word embedding is frozen during training. Following mBART, we use a language id symbol, i.e. `zh\_CN' or `de\_DE', as [EOS] of encoder inputs and [BOS] of decoder inputs for language identification. 

\noindent\textbf{Joint Training.}
Two independently pretrained networks are loaded as the initialization for joint training. Visual-language mapper (V-L Mapper) establishes a bridge between features of the visual modality and language modality. To reduce computational cost, we freeze the S3D backbone during joint training. For PHOENIX-2014T~\cite{camgoz2018neural} dataset, we use gloss representations (see Figure~3 of the paper) as the input of V-L Mapper. In CSL-Daily \cite{zhou2021improving} dataset, we observe that the gloss annotations almost contain all language information for generating the text, i.e., in many cases spoken text can be accurately predicted by simply reordering, or even copying, the gloss sequence. Therefore in CSL-Daily, the glosses' linguistic semantics are more helpful than their visual semantics for translation task (Sign2Text). As a consequence, we take the gloss probabilities (see Figure~3 of the paper) as the input of the V-L Mapper where the FC layer is initialized using the weights of the pretrained word embedding in the translation network. The other settings for the two datasets are identical. The learning rate is set as 1e-5 for the translation network and 1e-3 for trainable layers in the visual encoder and V-L mapper. We train the whole network under the joint supervision of the CTC loss and cross-entropy loss with a loss weight of 1.0 for 40 epochs. During evaluation, we use beam search decoding with a beam width of 4 and length penalty of 1.

\section{More ablations}
\noindent\textbf{Loss weights.} In Sign2Text joint training, we vary weights of CTC loss and Cross-Entropy loss to study their effects on translation performance. The results are shown in table~\ref{tab:loss_weight}. We find that our method is insensitive to the loss weights.
\begin{table*}[t]
    \centering
        \begin{tabular}{c c| c c c c c |c c c c c }
    \toprule
      \multicolumn{2}{c|}{Loss weights} & \multicolumn{5}{c|}{Dev} & \multicolumn{5}{c}{Test} \\  
    CTC& translation & R& B1 &B2 & B3 & B4 & R& B1 &B2 & B3 & B4 \\ \midrule
        0.5 & 1 & 52.95 & 54.03 & 41.32 & 33.23 & 27.70 & \textbf{53.17} & \textbf{54.49} & \textbf{42.01} & \textbf{33.90} & 28.33 \\
        1 & 1 & 53.10 & 53.95 & 41.12 & 33.14 & 27.61 & 52.65 & 53.97 &41.75 & 33.84 & \textbf{28.39} \\
        2 & 1 & 53.15 & 53.90 & 41.03 & 32.94 & 27.35 & 52.80 & 54.08 & 41.48 & 33.34 & 27.70 \\
        1 & 0.5 & 52.91 & 54.09 & 41.46 & \textbf{33.45} & 27.91 & 53.12 & 54.19 & 41.60 & 33.67 & 28.27 \\
        1 & 2 & \textbf{53.59} & \textbf{54.35} & \textbf{41.50} & 33.44 & \textbf{27.92} & 52.73 & 53.68 & 41.31 & 33.42 & 28.01 \\
     \bottomrule
    \end{tabular}
    \caption{Ablation study of varied weights on the CTC loss and translation loss on \textbf{PHOENIX Sign2Text} training. }
    \label{tab:loss_weight}
\end{table*}

\noindent\textbf{Temporal downsampling.} We also temporally downsample input videos to 1/2 and 1/3 and use the downsampled videos to train Sign2Gloss and Sign2Text of PHOENIX-2014T. Comparison between different downsampling strides are shown in table~\ref{tab:downsample}. It can be seen that temporal downsampling greatly degrades performance in both Sign2Gloss and Sign2Text.
\begin{table*}[t]
    \centering
    \begin{tabular}{c  |c c  | c c c c c |c c c c c }
    \toprule
            \multicolumn{1}{c|}{\multirow{3}{*}{Downsample rate}} & \multicolumn{2}{c|}{Sign2Gloss} & \multicolumn{10}{c}{Sign2Text}  \\   \cmidrule{2-13}
             & \multicolumn{1}{c}{Dev} & \multicolumn{1}{c|}{Test} & \multicolumn{5}{c|}{Dev} & \multicolumn{5}{c}{Test} \\
            & WER & WER & R & B1 &B2 & B3 &B4 & R&B1&B2&B3&B4\\ \midrule
          1   & \textbf{21.90} & \textbf{22.45} & 53.10 & 53.95 & 41.12 & 33.14 & 27.61 & \textbf{52.64} & \textbf{53.97} & \textbf{41.75} & \textbf{33.84} & \textbf{28.39} \\
          1/2  & 29.54 & 30.73 & 50.74 & 51.33 & 38.51 & 30.58 & 25.24 & 50.78 & 52.01 & 39.14 & 31.02 & 25.59 \\
          1/3  & 40.97 & 40.43 & 46.65 & 47.62 & 34.16 & 26.23 & 21.07 & 46.45 & 47.66 & 34.72 & 26.92 & 21.89 \\
     \bottomrule
    \end{tabular}
    \caption{Ablation study of different temporal downsampling rate temporal of input videos on PHOENIX dataset.}
    \label{tab:downsample}
\end{table*}

\begin{table*}[t!]
    \centering
        \begin{tabular}{c| c c c c c |c c c c c }
    \toprule
      \multirow{2}{*}{Frozen blocks} & \multicolumn{5}{c|}{Dev} & \multicolumn{5}{c}{Test} \\  
     & R& B1 &B2 & B3 & B4 & R& B1 &B2 & B3 & B4 \\ \midrule
     1   & \textbf{53.28} & 53.75 &41.14 & 33.31 & 27.95 & 52.35 & 53.57 &41.21 & 33.23 & 27.79 \\
     1-2   & 52.67 &53.42 &40.76 & 32.91 &  27.51 & 52.39 &53.57 &41.20 & 33.26 &  27.82 \\
     1-3    & 53.03 &\textbf{54.24} &\textbf{41.71} & \textbf{33.69} &  \textbf{28.16} & 52.61 &53.89 &41.48 & 33.51 &  28.03 \\
     1-4 (default)    & 53.10& 53.95 & 41.12 & 33.14 & 27.61 & \textbf{52.65} & \textbf{53.97} & \textbf{41.75} & \textbf{33.84} &\textbf{28.39}  \\  
    
     \bottomrule
    \end{tabular}
    \caption{Freezing different S3D blocks on \textbf{PHOENIX Sign2Text}.}
    \label{tab:end2end}
\end{table*}
\noindent\textbf{Tuning S3D layers in Sign2Text joint training}
We freeze the S3D backbone for efficient Sign2Text training. Table~\ref{tab:end2end} examines freezing a subset of S3D blocks. Tuning S3D layers does not improve the performance, for which we conjecture three reasons. 1) the hyperparameters for joint training with tunable S3D layers need re-tuning. 2) Excessive number of trainable parameters lead to overfitting. 3) With a well-pretrained visual encoder,  visual features are not the bottleneck for translation performance in Sign2Text join training.

\section{Qualitative Analysis}

We report some qualitative results on PHOENIX-2014T~\cite{camgoz2018neural} and CSL-Daily~\cite{zhou2021improving} datasets. We first demonstrate the effectiveness of our Sign2Text end-to-end training. Then we reveal the current method's limitations to shed some light on future work.
\paragraph{Effectiveness of Joint Sign2Text Training.}
To illustrate that our end-to-end Sign2Text training can utilize rich visual information from sign videos and semantic knowledge from text transcriptions to produce translation of high quality, we compare it with our Gloss2Text model and Sign2Gloss2Text pipeline. Table~\ref{tab:examples_good} shows the ground-truth glosses (Gloss), ground-truth text references (Text), gloss predictions from visual encoder (Sign2Gloss), and translation results from three approaches, namely Gloss2Text, Sign2Gloss2Text and Sign2Text. 

In Example (a) and (c), we can see that when Sign2Gloss model predicts wrong glosses, the two-stage pipeline (Sign2Gloss2Text) will be influenced, resulting in the wrong translation texts. For instance, in Example (c), the sign of `Selfish' is wrongly predicted as `Happiness' by Sign2Gloss model, which further misleads Sign2Gloss2Text pipeline to generate the wrong translation `We cannot do happy things'. Nevertheless, our end-to-end Sign2Text model mitigates the error propagation issue, e.g., correctly generating `We cannot do selfish things'. Moreover, Example (b) and (d) demonstrate that our end-to-end Sign2Text model outperforms Gloss2Text translation model by correctly predicting the words which are not included in the gloss annotations. For example, in Example (b), our Sign2Text model manages to translate some subtle words such as `dominate' and `half', and in Example (d), unlike Gloss2Text model that translates according to the gloss's literal meaning, our Sign2Text model translates the sign of `Sugar' into `drink'. This suggests that our Sign2Text model is capable of leveraging visual information from sign videos and supplement the knowledge that the discrete glosses can not fully capture.

\paragraph{Limitations.}
Here we present some failure cases in Table~\ref{tab:examples_fail}. We observe that the our method has some difficulty in identifying numbers and location entities due to their low frequency in the training corpus. Also, when dealing with long inputs, the translation results may either leave out some information or be not fluent. We look forward to overcoming the difficulty by handling long-tailed sign distribution in future works. There are several limitations of our work. First, our current approach relies on continuous sentence-level gloss annotations. Although some existing SLT datasets provide gloss annotations, they are expensive to obtain. We hope to lift the need for manual gloss labels in the future. Second, both CSL-Daily and PHOENIX-2014T are recorded under constrained conditions with limited vocabulary and number of signers. All of the signings are interpreted from spoken language. While our model achieves good results in these two benchmarks, it is worth further studying its performance in wild scenarios with large vocabulary, diverse signers and conversational signings. As with broader social impact, it should be
noted that sign language modelling may result in increased surveillance of deaf communities. 

\begin{table*}[]
    \centering
    \begin{tabular}{c|c}
        \specialrule{1.2pt}{3pt}{3pt}
        \multicolumn{2}{l}{\textbf{Example (a)}} \\
        \multirow{2}{*}{Gloss (GT)} & HEUTE / NACHT / NORD / REGION / ELF / GRAD  / ALPEN / EINS / GRAD \\ 
         & (Today / Night / North / Region / Eleven / Degrees / \textbf{Alps} / One / Degree) \\
          \midrule
        \multirow{2}{*}{Sign2Gloss} & HEUTE / NACHT / NORD / ELF / GRAD / REGION / EINS / GRAD  \\ 
         & (Today / Night / North / Eleven / Degrees / \textbf{Region} / One / Degrees) \\
        \midrule
        \midrule
        \multirow{2}{*}{Text (GT)} & Heute Nacht elf Grad im Norden und ein Grad an den Alpen. \\ 
         & (Tonight eleven degrees in the north and one degree in the \textbf{Alps}.) \\
        \midrule
         \multirow{2}{*}{Gloss2Text} & Heute Nacht elf Grad an der Nordsee ein Grad an den Alpen. \\
          & (Tonight eleven degrees on the north sea one degree on the \textbf{Alps}.) \\
        \midrule
         \multirow{2}{*}{Sign2Gloss2Text} & Heute Nacht elf Grad an der Nordsee ein Grad im Breisgau. \\
          & (Tonight eleven degrees on the north sea one degree in \textbf{Breisgau}.) \\
          \midrule
         \multirow{2}{*}{Sign2Text} & Heute Nacht elf Grad im Norden und ein Grad an den Alpen.  \\
          & (Tonight eleven degrees in the north and one degree in the \textbf{Alps}.) \\
          
        \specialrule{1.2pt}{3pt}{3pt}
        
        \multicolumn{2}{l}{\textbf{Example (b)}} \\
        \multirow{2}{*}{Gloss (GT)} & MORGEN / IX / NORDOST / WOLKE / ÜBERWIEGEND / SCHAUER \\ 
         & (Tomorrow / IX / Northeast / Clouds / Presently / Showers) \\
        \midrule
        \multirow{2}{*}{Sign2Gloss} & MORGEN / NORDOST / WOLKE / ÜBERWIEGEND / SCHAUER  \\ 
         & (Tomorrow / Northeast / Clouds / Presently / Showers) \\
        \midrule
        \midrule
        \multirow{2}{*}{Text (GT)} & Morgen überwiegen in der Nordosthälfte die Wolken und es fallen einzelne Schauer. \\ 
         & (Tomorrow the clouds will \textbf{dominate} in the northeast \textbf{half} and there will be a few showers.) \\
        \midrule
         \multirow{2}{*}{Gloss2Text} & Morgen ist es im Norden und Nordosten wolkig mit einzelnen Schauern. \\
          & (Tomorrow it will be cloudy in the north and northeast with a few showers.) \\
          \midrule
         \multirow{2}{*}{Sign2Gloss2Text} & Morgen im Norden und Nordosten dichte Wolken mit einzelnen Schauern. \\
          & (Tomorrow in the north and northeast dense clouds with a few showers.) \\
          \midrule
         \multirow{2}{*}{Sign2Text} & Morgen überwiegen in der Nordosthälfte die Wolken und es fallen einige Schauer.  \\
          & (Tomorrow the clouds will \textbf{dominate} in the northeast \textbf{half} and some showers will fall.) \\ 
          
        \specialrule{1.2pt}{3pt}{3pt}
        \multicolumn{2}{l}{\textbf{Example (c)}} \\
        Gloss (GT) & \begin{CJK*}{UTF8}{gbsn}我们 \ / 做 \ / 人 \ / 自私 \ / 不行\end{CJK*}  \ (We / Do / People / \textbf{Selfish} / Cannot) \\
        \midrule
         Sign2Gloss & \begin{CJK*}{UTF8}{gbsn}我们 \ / 做 \ / 幸福 \ / 不行\end{CJK*} \ (We / Do / \textbf{Happiness} / Cannot) \\ 
          \midrule
          \midrule
        Text (GT) & \begin{CJK*}{UTF8}{gbsn}我们做人不能自私。\end{CJK*} (We cannot be \textbf{selfish}.)\\%我们做人不能自私 \\
        \midrule
         Gloss2Text &\begin{CJK*}{UTF8}{gbsn}我们做人不能自私自利。\end{CJK*} (We cannot be \textbf{selfish}.) \\
          \midrule
         Sign2Gloss2Text & \begin{CJK*}{UTF8}{gbsn}我们不能做幸福的事情。\end{CJK*} (We cannot do \textbf{happy} things.) \\
          \midrule
         Sign2Text & \begin{CJK*}{UTF8}{gbsn}我们不能做自私自利的事情。\end{CJK*} (We cannot do \textbf{selfish} things.) \\
        \specialrule{1.2pt}{3pt}{3pt}

        \multicolumn{2}{l}{\textbf{Example (d)}} \\
        Gloss (GT) & \begin{CJK*}{UTF8}{gbsn}今天 \ / 菜 \ / 咸 \ / 我 \ / 想 \ / 喝 \ / 糖\end{CJK*} \ (Today / Food / Salty / I / Want to / Drink / \textbf{Sugar}) \\
          \midrule
         Sign2Gloss & \begin{CJK*}{UTF8}{gbsn}今天 \ / 菜 \ /  \ / 我 \ / 想 \ / 喝 \ / 糖\end{CJK*}  \ (Today / Food / Salty / I / Want to / Drink / \textbf{Sugar}) \\ 
        \midrule
        \midrule
        Text (GT)& \begin{CJK*}{UTF8}{gbsn}今天的菜好咸，我想喝饮料。\end{CJK*} \ (The food is quite salty today and I want to drink a \textbf{drink}.)\\
        \midrule
         Gloss2Text &\begin{CJK*}{UTF8}{gbsn}今天菜很咸,我想喝糖。\end{CJK*} \ (The food is quite salty today and I want to drink some \textbf{sugar}.) \\ 
          \midrule
         Sign2Gloss2Text & \begin{CJK*}{UTF8}{gbsn}今天菜很咸,我想喝糖。\end{CJK*} (The food is quite salty today and I want to drink some \textbf{sugar}.) \\
          \midrule
         Sign2Text & \begin{CJK*}{UTF8}{gbsn}今天的菜真咸,我想喝饮料。\end{CJK*} (The food is quite salty today and I want to drink a \textbf{drink}.) \\  
         \specialrule{1.2pt}{3pt}{3pt} 
         
    \end{tabular}
    \caption{Qualitative analysis on PHOENIX-2014T (Example (a) and (b)) and CSL-Daily (Example (c) and (d)) datasets.} 
    \label{tab:examples_good}
\end{table*}
\begin{table*}[t!]
    %\centering
    \begin{tabular}{c|c}
    \specialrule{1.2pt}{3pt}{3pt}
    \multicolumn{2}{l}{\textbf{Example (a)}} \\
        \multirow{2}{*}{Text (GT)} & Auch am Sonntag ist es neben Wolkenfeldern teilweise freundlich und trocken bei minus fünf bis plus zwei Grad. \\ & (On sunday, too, it is partly friendly and dry in addition to cloud fields, with minus five to plus \textbf{two} degrees.) \\
        \midrule
        \multirow{2}{*}{Sign2Text} & Am Sonntag teils wolkig teils freundlich und trocken minus fünf bis plus drei Grad. \\ &  (On sunday partly cloudy partly friendly and dry minus five to plus \textbf{three} degrees.)\\
         \specialrule{1.2pt}{3pt}{3pt} 
        \multicolumn{2}{l}{\textbf{Example (b)}} \\
        \multirow{2}{*}{Text (GT)} & Also wir haben morgen HöchstTemperaturen in Mitteleuropa von achtzehn bis sechsundzwanzig Grad. \\ & (So tomorrow we have maximum temperatures of \textbf{eighteen} to twenty-six degrees in central Europe.) \\
        \midrule
        \multirow{2}{*}{Sign2Text} & Morgen Temperaturen von fünfzehn Grad in Mitteleuropa bis sechsundzwanzig Grad in der Lausitz. \\ & (
Tomorrow temperatures from \textbf{fifteen} degrees in central Europe to twenty-six degrees in \textbf{lausitz}.)\\
        \specialrule{1.2pt}{3pt}{3pt}
        \multicolumn{2}{l}{\textbf{Example (c)}} \\
        \multirow{2}{*}{Text (GT)} & \begin{CJK*}{UTF8}{gbsn}他今年四岁。\end{CJK*} \\
        & (He is \textbf{four} years old.) \\
        \midrule
         \multirow{2}{*}{Sign2Text} & \begin{CJK*}{UTF8}{gbsn}他今年三岁 。\end{CJK*} \\
         &(He is \textbf{three} years old.)\\  
        \specialrule{1.2pt}{3pt}{3pt}
        \multicolumn{2}{l}{\textbf{Example (d)}} \\
        \multirow{2}{*}{Text (GT)} & \begin{CJK*}{UTF8}{gbsn}我们不但要从成功中总结经验，还要从失败中吸取教训。\end{CJK*} \\ & (
We must not only draw lessons from our success, but also learn from our failures.) \\
        \midrule
        \multirow{2}{*}{Sign2Text} & \begin{CJK*}{UTF8}{gbsn}我们要善于吸取失败的教训。 \end{CJK*} \\ & (
We must be good at learning from our failures.)\\
        \specialrule{1.2pt}{3pt}{3pt}
      
    \end{tabular}
    \caption{Failure cases from PHOENIX-2014T (Example (a) and (b)) and CSL-Daily (Example (c) and (d)) datasets. Our method has difficulty in recognizing numbers and location entities due to their low frequency in the training corpus.}
    \label{tab:examples_fail}
\end{table*}

% \input{sections/table}

%%%%%%%%% REFERENCES
%\newpage
{\small
\bibliographystyle{ieee_fullname}
\bibliography{egbib}

\begin{thebibliography}{10}\itemsep=-1pt

\bibitem{MASS}
Roee Aharoni, Melvin Johnson, and Orhan Firat.
\newblock Massively multilingual neural machine translation.
\newblock In Jill Burstein, Christy Doran, and Thamar Solorio, editors, {\em
  Proceedings of the 2019 Conference of the North American Chapter of the
  Association for Computational Linguistics: Human Language Technologies,
  {NAACL-HLT})}, 2019.

\bibitem{Albanie2020bsl1k}
Samuel Albanie, G{\"u}l Varol, Liliane Momeni, Triantafyllos Afouras, Joon~Son
  Chung, Neil Fox, and Andrew Zisserman.
\newblock {BSL-1K}: {S}caling up co-articulated sign language recognition using
  mouthing cues.
\newblock In {\em European Conference on Computer Vision}, 2020.

\bibitem{baziotis2020languageprior}
Christos Baziotis, Barry Haddow, and Alexandra Birch.
\newblock Language model prior for low-resource neural machine translation.
\newblock In {\em Proceedings of the Conference on Empirical Methods in Natural
  Language Processing (EMNLP)}, 2020.

\bibitem{camgoz2018neural}
Necati~Cihan Camgoz, Simon Hadfield, Oscar Koller, Hermann Ney, and Richard
  Bowden.
\newblock Neural sign language translation.
\newblock In {\em Proceedings of the IEEE Conference on Computer Vision and
  Pattern Recognition}, 2018.

\bibitem{Camg2020Multichannel}
Necati~Cihan Camg{\"{o}}z, Oscar Koller, Simon Hadfield, and Richard Bowden.
\newblock Multi-channel transformers for multi-articulatory sign language
  translation.
\newblock In {\em {ECCV} 2020 Workshops}, 2020.

\bibitem{camgoz2020sign}
Necati~Cihan Camgoz, Oscar Koller, Simon Hadfield, and Richard Bowden.
\newblock Sign language transformers: Joint end-to-end sign language
  recognition and translation.
\newblock In {\em Proceedings of the IEEE/CVF conference on computer vision and
  pattern recognition}, 2020.

\bibitem{carreira2017quo}
Joao Carreira and Andrew Zisserman.
\newblock Quo vadis, action recognition? a new model and the kinetics dataset.
\newblock In {\em proceedings of the IEEE Conference on Computer Vision and
  Pattern Recognition}, 2017.

\bibitem{xlm-r}
Alexis Conneau, Kartikay Khandelwal, Naman Goyal, Vishrav Chaudhary, Guillaume
  Wenzek, Francisco Guzm{\'a}n, Edouard Grave, Myle Ott, Luke Zettlemoyer, and
  Veselin Stoyanov.
\newblock Unsupervised cross-lingual representation learning at scale.
\newblock In {\em Proceedings of the 58th Annual Meeting of the Association for
  Computational Linguistics}, 2020.

\bibitem{Cui2017Recurrent}
Runpeng Cui, Hu Liu, and Changshui Zhang.
\newblock Recurrent convolutional neural networks for continuous sign language
  recognition by staged optimization.
\newblock In {\em IEEE Conference on Computer Vision and Pattern Recognition
  (CVPR)}, 2017.

\bibitem{BERT}
Jacob Devlin, Ming{-}Wei Chang, Kenton Lee, and Kristina Toutanova.
\newblock {BERT:} pre-training of deep bidirectional transformers for language
  understanding.
\newblock In {\em Proceedings of the Conference of the North American Chapter
  of the Association for Computational Linguistics: Human Language
  Technologies, {NAACL-HLT}}, 2019.

\bibitem{dosovitskiy2020vit}
Alexey Dosovitskiy, Lucas Beyer, Alexander Kolesnikov, Dirk Weissenborn,
  Xiaohua Zhai, Thomas Unterthiner, Mostafa Dehghani, Matthias Minderer, Georg
  Heigold, Sylvain Gelly, Jakob Uszkoreit, and Neil Houlsby.
\newblock An image is worth 16x16 words: Transformers for image recognition at
  scale.
\newblock {\em ICLR}, 2021.

\bibitem{Duarte2021how2}
Amanda Duarte, Shruti Palaskar, Lucas Ventura, Deepti Ghadiyaram, Kenneth
  DeHaan, Florian Metze, Jordi Torres, and Xavier Giro-i Nieto.
\newblock How2sign: A large-scale multimodal dataset for continuous american
  sign language.
\newblock In {\em Proceedings of the IEEE/CVF Conference on Computer Vision and
  Pattern Recognition (CVPR)}, 2021.

\bibitem{fan2020pyslowfast}
Haoqi Fan, Yanghao Li, Bo Xiong, Wan-Yen Lo, and Christoph Feichtenhofer.
\newblock Pyslowfast.
\newblock \url{https://github.com/facebookresearch/slowfast}, 2020.

\bibitem{Feichtenhofer2020X3D}
Christoph Feichtenhofer.
\newblock X3d: Expanding architectures for efficient video recognition.
\newblock In {\em Proceedings of the IEEE/CVF Conference on Computer Vision and
  Pattern Recognition (CVPR)}, 2020.

\bibitem{Raghav2017Sth}
Raghav Goyal, Samira~Ebrahimi Kahou, Vincent Michalski, Joanna Materzynska,
  Susanne Westphal, Heuna Kim, Valentin Haenel, Ingo Fr{\"{u}}nd, Peter
  Yianilos, Moritz Mueller{-}Freitag, Florian Hoppe, Christian Thurau, Ingo
  Bax, and Roland Memisevic.
\newblock The "something something" video database for learning and evaluating
  visual common sense.
\newblock In {\em {IEEE} International Conference on Computer Vision, {ICCV}},
  2017.

\bibitem{graves2006connectionist}
Alex Graves, Santiago Fern{\'a}ndez, Faustino Gomez, and J{\"u}rgen
  Schmidhuber.
\newblock Connectionist temporal classification: labelling unsegmented sequence
  data with recurrent neural networks.
\newblock In {\em Proceedings of the 23rd international conference on Machine
  learning}, 2006.

\bibitem{kaiming2016cvpr}
Kaiming He, Xiangyu Zhang, Shaoqing Ren, and Jian Sun.
\newblock Deep residual learning for image recognition.
\newblock In {\em {IEEE} Conference on Computer Vision and Pattern
  Recognition}, 2016.

\bibitem{Fabian2015Act}
Fabian~Caba Heilbron, Victor Escorcia, Bernard Ghanem, and Juan~Carlos Niebles.
\newblock Activitynet: {A} large-scale video benchmark for human activity
  understanding.
\newblock In {\em {IEEE} Conference on Computer Vision and Pattern Recognition,
  {CVPR}}, 2015.

\bibitem{imashev2020krsl}
Alfarabi Imashev, Medet Mukushev, Vadim Kimmelman, and Anara Sandygulova.
\newblock A dataset for linguistic understanding, visual evaluation, and
  recognition of sign languages: The k-{RSL}.
\newblock In {\em Proceedings of the 24th Conference on Computational Natural
  Language Learning}, 2020.

\bibitem{MSASL}
Hamid Reza~Vaezi Joze and Oscar Koller.
\newblock {MS-ASL:} {A} large-scale data set and benchmark for understanding
  american sign language.
\newblock In {\em 30th British Machine Vision Conference, {BMVC}}, 2019.

\bibitem{K400_dataset}
Will Kay, Jo{\~{a}}o Carreira, Karen Simonyan, Brian Zhang, Chloe Hillier,
  Sudheendra Vijayanarasimhan, Fabio Viola, Tim Green, Trevor Back, Paul
  Natsev, Mustafa Suleyman, and Andrew Zisserman.
\newblock The kinetics human action video dataset.
\newblock {\em CoRR}, 2017.

\bibitem{koller2019weak}
Oscar Koller, Necati~Cihan Camgoz, Hermann Ney, and Richard Bowden.
\newblock Weakly supervised learning with multi-stream cnn-lstm-hmms to
  discover sequential parallelism in sign language videos.
\newblock {\em IEEE Transactions on Pattern Analysis and Machine Intelligence},
  2019.

\bibitem{koller2017re-sign}
Oscar Koller, Sepehr Zargaran, and Hermann Ney.
\newblock Re-sign: Re-aligned end-to-end sequence modelling with deep recurrent
  cnn-hmms.
\newblock In {\em IEEE Conference on Computer Vision and Pattern Recognition},
  2017.

\bibitem{SentencePiece}
Taku Kudo and John Richardson.
\newblock Sentencepiece: {A} simple and language independent subword tokenizer
  and detokenizer for neural text processing.
\newblock In Eduardo Blanco and Wei Lu, editors, {\em Proceedings of the 2018
  Conference on Empirical Methods in Natural Language Processing, {EMNLP}:
  System Demonstrations}, 2018.

\bibitem{XLM}
Guillaume Lample and Alexis Conneau.
\newblock Cross-lingual language model pretraining.
\newblock {\em Advances in Neural Information Processing Systems (NeurIPS)},
  2019.

\bibitem{lei2021less}
Jie Lei, Linjie Li, Luowei Zhou, Zhe Gan, Tamara~L. Berg, Mohit Bansal, and
  Jingjing Liu.
\newblock Less is more: Clipbert for video-and-language learningvia sparse
  sampling.
\newblock In {\em CVPR}, 2021.

\bibitem{Mike2019BART}
Mike Lewis, Yinhan Liu, Naman Goyal, Marjan Ghazvininejad, Abdelrahman Mohamed,
  Omer Levy, Veselin Stoyanov, and Luke Zettlemoyer.
\newblock {BART:} denoising sequence-to-sequence pre-training for natural
  language generation, translation, and comprehension.
\newblock In {\em Proceedings of the 58th Annual Meeting of the Association for
  Computational Linguistics}, 2020.

\bibitem{li2020word}
Dongxu Li, Cristian Rodriguez, Xin Yu, and Hongdong Li.
\newblock Word-level deep sign language recognition from video: A new
  large-scale dataset and methods comparison.
\newblock In {\em The IEEE Winter Conference on Applications of Computer
  Vision}, 2020.

\bibitem{li2020tspnet}
Dongxu Li, Chenchen Xu, Xin Yu, Kaihao Zhang, Benjamin Swift, Hanna Suominen,
  and Hongdong Li.
\newblock Tspnet: Hierarchical feature learning via temporal semantic pyramid
  for sign language translation.
\newblock In {\em Advances in Neural Information Processing Systems}, 2020.

\bibitem{li2020transferring}
Dongxu Li, Xin Yu, Chenchen Xu, Lars Petersson, and Hongdong Li.
\newblock Transferring cross-domain knowledge for video sign language
  recognition.
\newblock In {\em Proceedings of the IEEE/CVF Conference on Computer Vision and
  Pattern Recognition}, 2020.

\bibitem{li2020oscar}
Xiujun Li, Xi Yin, Chunyuan Li, Xiaowei Hu, Pengchuan Zhang, Lei Zhang, Lijuan
  Wang, Houdong Hu, Li Dong, Furu Wei, Yejin Choi, and Jianfeng Gao.
\newblock Oscar: Object-semantics aligned pre-training for vision-language
  tasks.
\newblock {\em ECCV}, 2020.

\bibitem{rouge}
Chin-Yew Lin.
\newblock {ROUGE}: A package for automatic evaluation of summaries.
\newblock In {\em Text Summarization Branches Out}, pages 74--81, Barcelona,
  Spain, July 2004. Association for Computational Linguistics.

\bibitem{liu2020multilingual}
Yinhan Liu, Jiatao Gu, Naman Goyal, Xian Li, Sergey Edunov, Marjan
  Ghazvininejad, Mike Lewis, and Luke Zettlemoyer.
\newblock Multilingual denoising pre-training for neural machine translation.
\newblock {\em Transactions of the Association for Computational Linguistics},
  2020.

\bibitem{liu2021Swin}
Ze Liu, Yutong Lin, Yue Cao, Han Hu, Yixuan Wei, Zheng Zhang, Stephen Lin, and
  Baining Guo.
\newblock Swin transformer: Hierarchical vision transformer using shifted
  windows.
\newblock {\em International Conference on Computer Vision (ICCV)}, 2021.

\bibitem{Luo2020UniVL}
Huaishao Luo, Lei Ji, Botian Shi, Haoyang Huang, Nan Duan, Tianrui Li, Jason
  Li, Taroon Bharti, and Ming Zhou.
\newblock Univl: A unified video and language pre-training model for multimodal
  understanding and generation.
\newblock {\em arXiv preprint arXiv:2002.06353}, 2020.

\bibitem{bleu}
Kishore Papineni, Salim Roukos, Todd Ward, and Wei{-}Jing Zhu.
\newblock Bleu: a method for automatic evaluation of machine translation.
\newblock In {\em Proceedings of the 40th Annual Meeting of the Association for
  Computational Linguistics}, 2002.

\bibitem{Pu2019Iterative}
Junfu Pu, Wengang Zhou, and Houqiang Li.
\newblock Iterative alignment network for continuous sign language recognition.
\newblock In {\em {IEEE} Conference on Computer Vision and Pattern Recognition,
  {CVPR}}, 2019.

\bibitem{qiu2017learning}
Zhaofan Qiu, Ting Yao, and Tao Mei.
\newblock Learning spatio-temporal representation with pseudo-3d residual
  networks.
\newblock In {\em {IEEE} International Conference on Computer Vision, {ICCV}},
  2017.

\bibitem{Radford@Clip}
Alec Radford, Jong~Wook Kim, Chris Hallacy, Aditya Ramesh, Gabriel Goh,
  Sandhini Agarwal, Girish Sastry, Amanda Askell, Pamela Mishkin, Jack Clark,
  Gretchen Krueger, and Ilya Sutskever.
\newblock Learning transferable visual models from natural language
  supervision.
\newblock In Marina Meila and Tong Zhang, editors, {\em Proceedings of the 38th
  International Conference on Machine Learning, {ICML}}, 2021.

\bibitem{Radford2019gpt}
Alec Radford, Jeff Wu, Rewon Child, David Luan, Dario Amodei, and Ilya
  Sutskever.
\newblock Language models are unsupervised multitask learners.
\newblock 2019.

\bibitem{Raffel2020T5}
Colin Raffel, Noam Shazeer, Adam Roberts, Katherine Lee, Sharan Narang, Michael
  Matena, Yanqi Zhou, Wei Li, and Peter~J. Liu.
\newblock Exploring the limits of transfer learning with a unified text-to-text
  transformer.
\newblock {\em Journal of Machine Learning Research}, 2020.

\bibitem{Sennrich2019Lowresource}
Rico Sennrich and Biao Zhang.
\newblock Revisiting low-resource neural machine translation: A case study.
\newblock In {\em Proceedings of the 57th Annual Meeting of the Association for
  Computational Linguistics}, 2019.

\bibitem{autsl}
Ozge~Mercanoglu Sincan and Hacer~Yalim Keles.
\newblock Autsl: A large scale multi-modal turkish sign language dataset and
  baseline methods.
\newblock {\em IEEE Access}, 2020.

\bibitem{vaswani2017attention}
Ashish Vaswani, Noam Shazeer, Niki Parmar, Jakob Uszkoreit, Llion Jones,
  Aidan~N Gomez, {\L}ukasz Kaiser, and Illia Polosukhin.
\newblock Attention is all you need.
\newblock In {\em Advances in neural information processing systems}, 2017.

\bibitem{xie2018rethinking}
Saining Xie, Chen Sun, Jonathan Huang, Zhuowen Tu, and Kevin Murphy.
\newblock Rethinking spatiotemporal feature learning: Speed-accuracy trade-offs
  in video classification.
\newblock In {\em Proceedings of the European conference on computer vision
  (ECCV)}, 2018.

\bibitem{Yin2020STMCTransf}
Kayo Yin and Jesse Read.
\newblock Better sign language translation with {STMC}-transformer.
\newblock In {\em Proceedings of the 28th International Conference on
  Computational Linguistics}, 2020.

\bibitem{zhang2021vinvl}
Pengchuan Zhang, Xiujun Li, Xiaowei Hu, Jianwei Yang, Lei Zhang, Lijuan Wang,
  Yejin Choi, and Jianfeng Gao.
\newblock Vinvl: Making visual representations matter in vision-language
  models.
\newblock {\em CVPR}, 2021.

\bibitem{VinVL_2021}
Pengchuan Zhang, Xiujun Li, Xiaowei Hu, Jianwei Yang, Lei Zhang, Lijuan Wang,
  Yejin Choi, and Jianfeng Gao.
\newblock Vinvl: Revisiting visual representations in vision-language models.
\newblock In {\em {IEEE} Conference on Computer Vision and Pattern Recognition,
  {CVPR}}, 2021.

\bibitem{zhou2021improving}
Hao Zhou, Wengang Zhou, Weizhen Qi, Junfu Pu, and Houqiang Li.
\newblock Improving sign language translation with monolingual data by sign
  back-translation.
\newblock In {\em Proceedings of the IEEE/CVF Conference on Computer Vision and
  Pattern Recognition}, 2021.

\bibitem{haozhou2020STMC}
Hao Zhou, Wengang Zhou, Yun Zhou, and Houqiang Li.
\newblock Spatial-temporal multi-cue network for sign language recognition and
  translation.
\newblock {\em IEEE Transactions on Multimedia}, 2022.

\end{thebibliography}
}

\end{document}